\documentclass{article}

\usepackage[final]{nips_2017}

\usepackage{color}
\usepackage[utf8]{inputenc} 
\usepackage[T1]{fontenc}    
\usepackage{hyperref}       
\usepackage{url}            
\usepackage{booktabs}       
\usepackage{amsfonts}       
\usepackage{nicefrac}       
\usepackage{microtype}      

\usepackage[tableposition=top]{caption}
\usepackage{times}
\usepackage{graphicx} 
\usepackage{subfigure} 
\usepackage{multirow}
\usepackage{microtype}      
\usepackage{cite}

\usepackage{algorithm}
\usepackage{algorithmic}
\usepackage{array}

\newcolumntype{L}{>{\centering\arraybackslash}m{1.7cm}}
\newcolumntype{M}{>{\centering\arraybackslash}m{3.2cm}}

\title{Interpreting Convolutional Neural Networks Through Compression}

\author{
  Reza Abbasi-Asl\\
  Department of Electrical Engineering\\
  and Computer Sciences,\\
  University of California, Berkeley,\\
  CA, 94720 \\
  \texttt{abbasi@berkeley.edu}
   \And
   Bin Yu \\
   Department of Statistics,\\
   Department of Electrical Engineering\\
   and Computer Sciences,\\
   University of California, Berkeley, \\
   CA, 94720 \\
   \texttt{binyu@berkeley.edu} \\
}

\begin{document}

\maketitle

\begin{abstract}

Convolutional neural networks (CNNs) achieve state-of-the-art performance in a wide variety of tasks in computer vision. However, interpreting CNNs still remains a challenge. This is mainly due to the large number of parameters in these networks. Here, we investigate the role of compression and particularly pruning filters in the interpretation of CNNs. We exploit our recently-proposed greedy structural compression scheme that prunes filters in a trained CNN. In our compression, the filter importance index is defined as the classification accuracy reduction (CAR) of the network after pruning that filter. The filters are then iteratively pruned based on the CAR index. We demonstrate the interpretability of CAR-compressed CNNs by showing that our algorithm prunes filters with visually redundant pattern selectivity. Specifically, we show the importance of shape-selective filters for object recognition, as opposed to color-selective filters. Out of top 20 CAR-pruned filters in AlexNet, 17 of them in the first layer and 14 of them in the second layer are color-selective filters. Finally, we introduce a variant of our CAR importance index that quantifies the importance of each image class to each CNN filter. We show that the most and the least important class labels present a meaningful interpretation of each filter that is consistent with the visualized pattern selectivity of that filter. 

\end{abstract}
\section{Structural compression }
Artificial deep neural networks (DNN) have achieved cutting-edge performance for many tasks in Artificial Intelligence such as machine learning, computer vision, autonomous driving and natural language and speech processing. However, the substantial number of weights in DNNs has limited their applications in devices such as smart phones and robots due to limited memory and computational power. Beside significant save in resources, compressed networks with less number of parameters are easier to be investigated or interpreted by humans for possible domain knowledge gain. To achieve this goal, a compression scheme requires to have at least two properties. First, the performance of the compressed network should be close to original uncompressed network. Second, the compression method should take into account the structure of the network. For example, considering a convolutional neural network (CNN), convolutional filters are the smallest meaningful components of a CNN. Therefore, to uncover redundant information in a network and build more interpretable model, it is natural to compress CNNs based on removing "less important" filters. We call such schemes "structural compression" schemes.

There have been many works on compressing deep neural networks. They mostly focus on reducing the number and size of the weights or parameters by pruning and quantizing them without considering the functionality of filters in each layer. Optimal brain damage \cite{lecun1989optimal}, optimal brain surgeon \cite{hassibi1993optimal}, Deep Compression \cite{han2015deep} and most recently SqueezeNet \cite{iandola2016squeezenet} are some examples. He et al. \cite{he2014reshaping} and Li et al. \cite{li2016pruning} have studied structural compression based on removing filters and introduced importance indexes based on average of incoming or outgoing weights to a filter. However, these importance measures typically do not yield satisfactory compressions of CNNs \cite{he2014reshaping} because of the substantially reduced classification accuracy as a result. 

Recently, we have introduced a set of structural compression rules for convolutional neural networks \cite{abbasi2017structural}. We have proposed an importance index for filters in CNNs. This index is based on the classification accuracy reduction (CAR) (or similarly for regression, RAR). A greedy structural compression scheme for CNNs is then proposed based on CAR. In our CAR structural (or filter pruning) compression algorithm, the filter with the least effect on the classification accuracy gets pruned in each iteration. The network can be retrained in each iteration and after pruning a filter. This process is regarded as \textit{fine tuning}. Our algorithm achieves state-of-the-art classification accuracy and compression rate among other structural compression schemes. Pruning half of the filters in either of the individual convolutional layers in AlexNet, our CAR algorithm achieves 16\% (for the layer 5) to 25\% (for the layer 2)  higher classification accuracies compared to the best benchmark filter pruning scheme (pruning based on average outgoing weights). If we retrain the pruned network, it achieves 50\% to 52\% classification accuracy (compared with 57\% of the uncompressed AlexNet). Similarly, to have a close-to-original classification accuracy (or 54\%) for both our and benchmark schemes, our CAR algorithm (without retraining) can achieve a compression ratio of around 1.2 (for layer 1) to 1.5 (for layer 5), which is 21\% to 43\% higher than those from the benchmark methods.

Here, we concentrate on the interpretability of a CAR compressed CNN. We show that through considering the network structure, our compression algorithm has the unique advantage of being more accessible to human interpreters. To our knowledge, we report the first extensive result on the connections between compression and interpretability of CNNs. 

\begin{figure}[!t]
  \begin{center}
  \centerline{\includegraphics[width=.9\columnwidth]{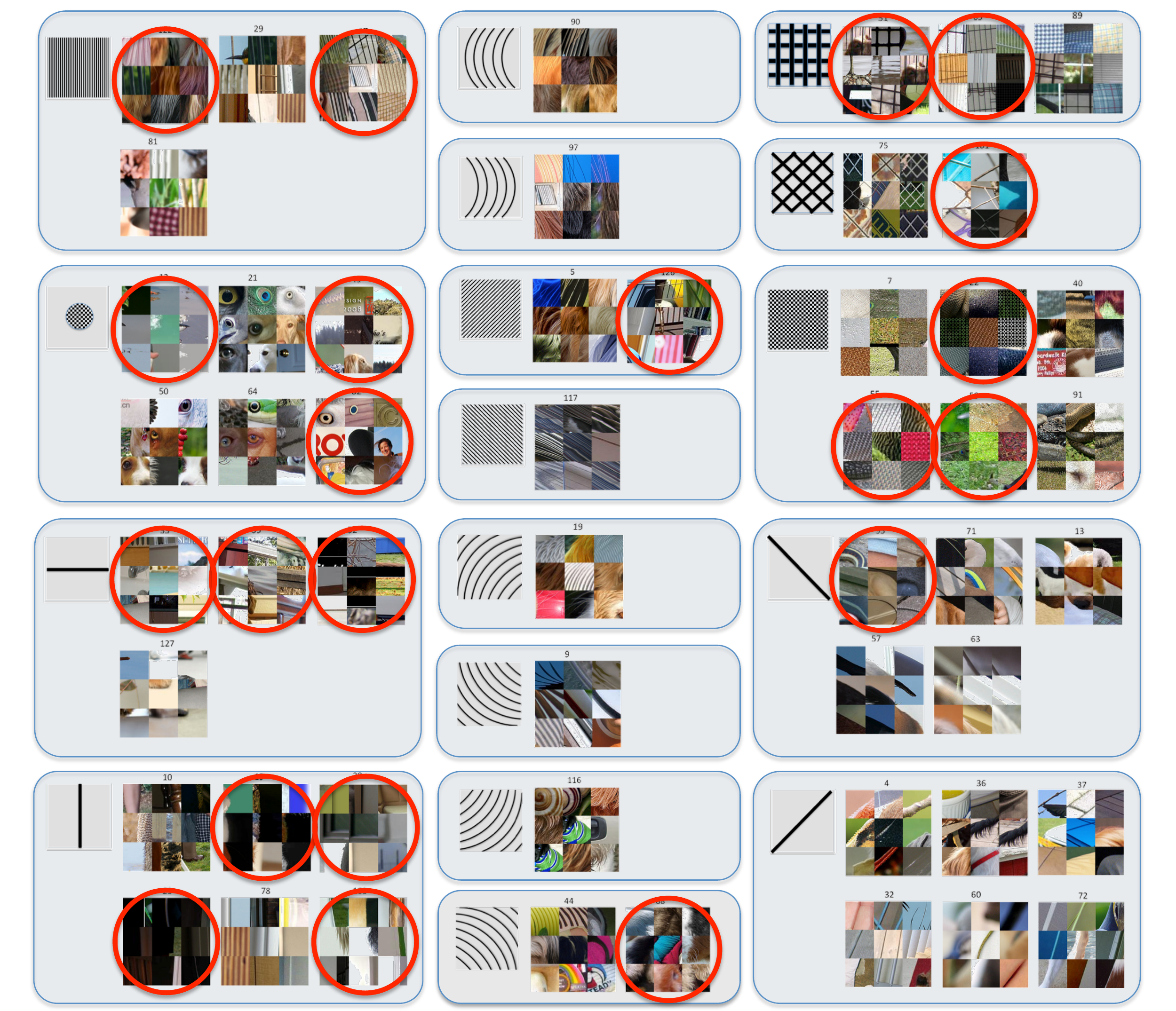}}
  \caption{CAR compression removes filters with visually redundant functionality from second layer of AlexNet. Each filter is visualized by the 9 image patches with highest responses for that filter. We have manually clustered 256 filters in the second layer of AlexNet into 30 groups (17 of them visualized here). Pattern selectivity of each group is illustrated in the left top corner of each box using a manually designed patch. We continue to iterate the CAR-based algorithm while the classification accuracy is in the range of relative 5\% from the accuracy of uncompressed network. This leads to pruning 102 filter from 256 filters in this layer. The pruned filters are specified with a red circle.}
  \label{redundent}
  \end{center}
\end{figure} 

\begin{figure}[!t]
		\begin{center}
		\centerline{\includegraphics[width=1\columnwidth]{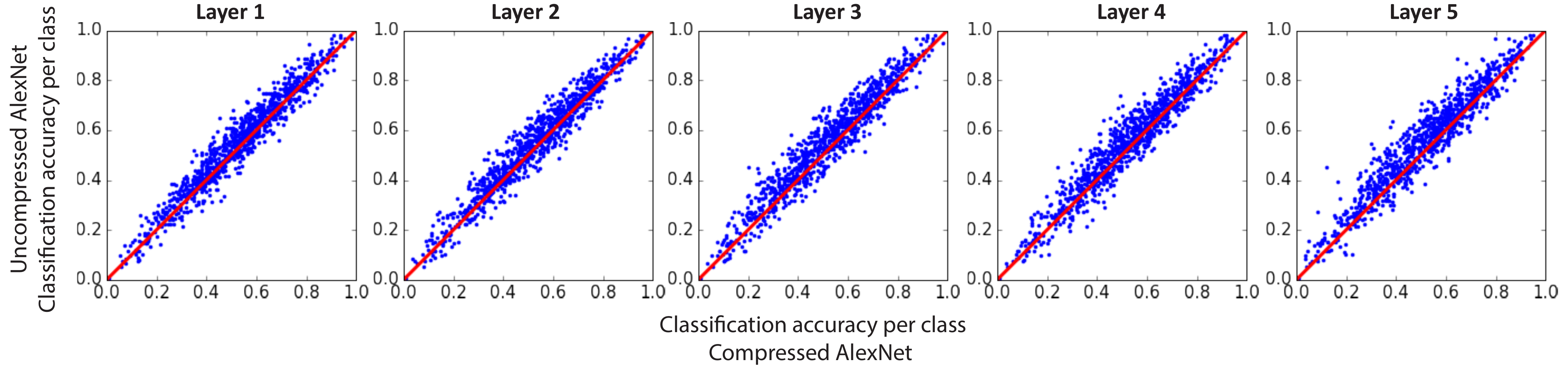}}
		\caption{Classification accuracy for each class of image in AlexNet after each individual layer is compressed compared to the uncompressed network. Each point in plots corresponds to one of the 1000 categories of images in test set.}
		\label{perclass}
		\end{center}
\end{figure}

\section{CAR compression prunes redundant filters}

To study the ability of the CAR compression method to identify redundant filters in CNNs, we take a closer look at the compressed networks. In this section, we focus on the second layer of AlexNet. It has 256 filters with visually diverse functionalities, which is considerably more than the numbers of filters in LeNet or the first layer of AlexNet. It is also easier to visualize filters in this layer compared to higher layers of AlexNet. However, similar results hold for the other layers of AlexNet and LeNet. We first perform CAR structural compression to prune 256 filters in the second layer of AlexNet, and continued to iterate the algorithm while the classification accuracy is 54\% or within a relative 5\% from the accuracy of uncompressed network. This led to pruning 102 filter from 256 filters in this layer. A subset of the removed and remaining filters are visualized in Figure \ref{redundent}. To visualize the pattern selectivity of each filter, we have fed one million image patch to the network and showed the top 9 image patch that activate a filter. This approach has been previously used to study functionality of filters in deep CNNs \cite{zeiler2014visualizing}. We have manually grouped filters with visually similar pattern selectivity (blue boxes in \ref{redundent}). A hand crafted image has been shown beside each group to demonstrate the function of filters in that group. Our algorithm tends to keep at least one filter from each group, suggesting that our greedy filter pruning process is able to identify redundant filters. This indicates that pruned filters based on the CAR importance index have in fact redundant functionality in the network. Looking deeper into the CAR indexes, out of top 20 pruned filters, 17 of them in the first layer and 14 of them in the second layer correspond to the color filters, respectively. This finding points to the fact that shape is often first-order important for object recognition. To further investigate the effect of compression of each of the convolutional layers, we have shown the scatter plots of classification accuracy for each of the 1000 classes in ImageNet in Figure \ref{perclass}. Although the total classification accuracy is about a relative 5\% lower for the each compressed network, the accuracies for many of the categories are comparable between compressed and uncompressed networks. In fact, 37\% (for layer 5) to 49\% (for layer 2) of the categories have accuracies no larger than 3\% below those for the uncompressed network. 
    
\section{Class-based interpretation of filters}
\begin{figure}[!t]
		\begin{center}
		\centerline{\includegraphics[width=1\columnwidth]{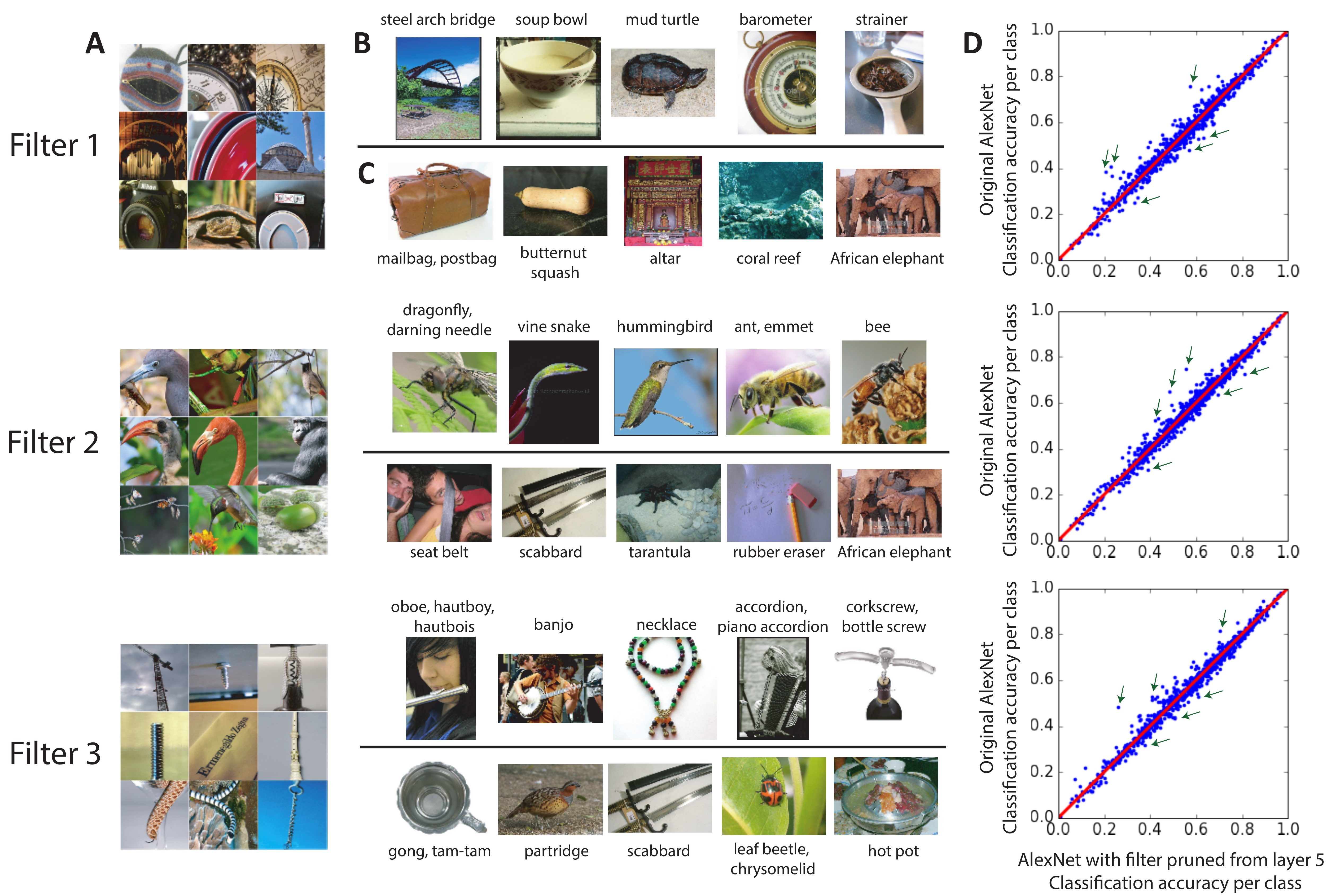}}
		\caption{The interpretation based on $CAR^c$ is consistent with the visualized pattern selectivity of each filter in layer 5 of AlexNet. Panel A shows The top 9 image patches that activate each filter \cite{zeiler2014visualizing}. Panel B and C show the top and bottom 5 classes with highest and lowest $CAR^c$, respectively. Besides the class label, one sample image from that class is also visualized. Panel D shows the scatter plot of classification accuracy for each of the 1000 classes in ImageNet. Three of the top and bottom classes with highest and lowest $CAR^c$ are pointed out with green arrows. Each row corresponds to one filter in layer 5 of AlexNet.}
		\label{perclass_interp}
		\end{center}
\end{figure}

With a slight modification in our definition for the CAR importance index, we build a new index that enables us to interpret the filters in CNNs with respect to image classes. We define $CAR^c(i,L)$ to be classification accuracy reduction in class $c$ when filter $i$ in layer $L$ is pruned. $CAR^c$ identifies the set of classes that their classification accuracy highly depends on the existence of a filter. These classes are the ones with highest $CAR^c$ among all of the classes. Similarly, for each filter, the performance in classes with the smallest $CAR^c$ have less dependency to that filter. Note that both CAR and  $CAR^c$ indexes could be negative numbers, that is the pruned network has higher classification accuracy compared to the original network. The labels of the two sets of classes with highest and lowest $CAR^c$ present a verbal interpretation of each filter in the network. $CAR^c$-based interpretation is a better fit for the higher layers in the CNN because filters in these layers are more abstract and therefore more explainable by the class labels. In  Figure \ref{perclass_interp} we show that the interpretation based on $CAR^c$ is consistent with the visualized pattern selectivity of each filter in layer 5 of AlexNet. Similar to previous section, the visualization is based on top 9 image patch activating that filter. We have selected three filters in layer 5 that are among the most important filters in this layer based on our original CAR pruning. Similar to the visualization in Figure \ref{perclass}, panel A illustrates the top 9 image patches that activate each filter. Panels B and C show the top and bottom 5 classes with highest and lowest $CAR^c$, respectively. Besides the class label, one sample image from that class is also visualized. Some of these classes are pointed out with green arrows in the scatter plot of classification accuracy for 1000 classes in ImageNet (panel D). The classes with highest $CAR^c$ share similar patterns visible in the top 9 patch activating each filter. For filter 1, the smooth elliptic curvature that consistently appears in the classes such as \textit{steep arch bridge} or \textit{soup bowel} is visible in the top activating patches (Panel A). On the other hand, less elliptic curvature patterns are expected in classes such as \textit{mailbag} or \textit{altar}. Filter 2 has higher $CAR^c$ for classes that contains patterns such as insect or bird's head. Filter 3 is mostly selected by the classes that contain images of a single long tool, particularly musical instruments such as \textit{oboe} or \textit{banjo}.

\section{Concluding remarks}
Structural compression (or filter pruning) of CNNs has the dual purposes of saving memory and computational cost on small devices, and of resulted CNNs being more humanly interpretable. We showed that the CAR-compression prunes filters with visually redundant functionalities. We also proposed a variant of CAR index to compare classification accuracies of original and pruned CNNs for each image class. In general, a similar comparison can be carried out for any two networks through our importance index. This is a fruitful direction to pursue, particularly given the recent wave of various CNNs with different structures. Finally, we expect that our CAR structural compression algorithm for CNNs and related interpretations can be adapted to fully-connected networks with modifications.

\section*{Acknowledgements}
This work is supported by National Science Foundation (NSF) Grant DMS-1613002, and the Center for Science of Information, an NSF Science and Technology Center, under Grant Agreement CCF-0939370. We gratefully acknowledge the support of NVIDIA Corporation with the donation of the Tesla GPU used for this research.


\end{document}